# Development and Evaluation of a Personalized Computer-aided Question Generation for English Learners to Improve Proficiency and Correct Mistakes

Yi-Ting Huang, Meng Chang Chen, and Yeali S. Sun

**Abstract**—In the last several years, the field of computer–assisted language learning has increasingly focused on computer–aided question generation. However, this approach often provides examinees with an exhaustive amount of questions that are not designed for any specific testing purpose. In this work, we present a personalized computer-aided question generation that generates multiple–choice questions at various difficulty levels and types, including vocabulary, grammar and reading comprehension. In order to improve the weaknesses of examinees, it selects questions depending on a student's estimated proficiency level and unclear concepts behind incorrect responses. This study's results show that the students with the personalized automatic quiz generation corrected their mistakes more frequently than ones only with computer–aided question generation. Moreover, students demonstrated the most progress between the pre–test and post–test and correctly answered more difficult questions. Finally, we investigated the personalizing strategy and found that a student could make a significant progress if the proposed system offered the vocabulary questions at the same level of a student's proficiency level, and if the grammar and reading comprehension questions were at a level lower than the student's.

———————————— ◆ ————————————

## 1 INTRODUCTION

For several years, educational assessment has been playing an increasingly important role in teaching and learning [1], being used to evaluate the effectiveness of teaching, diagnose the state of learning, and help the development of students' learning [2], [3], [4], [5]. With the development of computers and the Internet, Computer Adaptive Testing (CAT) is now developing a way to administer tests that adapt to learners' knowledge or competence in a language. Based on adaptive tests, it has been found that examinees' abilities can not only be more accurately measured using fewer suitable questions [6], [7], but also that student performance has been improved [5], [8]. Because CAT can provide both questions and also scaffolding hints and instructional feedback [9], student learning and knowledge acquisition is enhanced with external help. However, one of the problems of this approach is that a pool of items must be prepared and ready before applying the adaptive test. Given that a great number of potential reading materials continues to exponentially grow every day, there is room to overcome the major limitation of assessment resources: it is both time–consuming and cost–intensive for human experts to man-ually produce appropriate questions.

Recent years have seen increased attention given to the applications of Natural Language Processing (NLP) in the field of Computer–Assisted Language Learning (CALL), such as for grammar checkers [10], chat-bots [11] and computer–aided question generations [12], [13]. One of the more rapidly advancing subfields is computer-aided question generation, which generates questions automatically when a learning material is given. There is a multitude of studies now available for designing different question types, such as multiple–choice questions [14], [15], [16], [17], [18], [19], [20], [21], cloze tests [12], [22], [23], [24] and TOEFL synonym questions [25]. While these studies all investigate computer-aided question generation, there has been little research discussing the characteristic of questions during the procedure of the question generation process. Generally, very difficult questions can frustrate students, while overly-simplistic questions can demotivate them. Moreover, it is hard to create a meaningful test purpose while maximizing examinee learning outcomes because personalized design mechanisms [2], [3], [26] are still critically lacking.

In this work, we present a personalized automatic multiple-choice question generation for learning English as a foreign language. It can generate three question types, including vocabulary, grammar and reading comprehension, with different question difficulties. The term "personalized" refers to the adjustments made according to

———————————————————

- *Y.T. Huang is is with the Institute of Information Science, Academia Sinica, Taipei, Taiwan. E-mail: ythuang@iis.sinica.edu.tw.*
- *Meng Chang Chen is with the Institute of Information Science, Academia Sinica, Taipei, Taiwan. E-mail: mcc@iis.sinica.edu.tw.*
- *Yali S. Sun is with the the Department of Information Management, National Taiwan University, Taipei, Taiwan. E-mail: sunny@ntu.edu.tw.*

the students' needs, preferences or the context. In this study, we focus on student needs by matching the item difficulty levels to students' knowledge levels. For example, for two students with different levels, like a less-proficient student and a more-proficient student, read the same learning material about "apple". In our definition of personalized questions, questions should be selected according to item difficulty levels and the students' knowledge levels. That is, the less-proficient student should have a simple question while the high-proficient student should have a difficult question. In order to facilitate meaningful test purpose and maximize students' learning outcomes, the proposed system takes each student's knowledge level and a student's history record (incorrect answered question records) into consideration, and generates specific questions. Therefore, the research questions addressed in this study are:

1. Do the personalized computer-aided questions help students correct their previous mistakes?
2. Do students who answer personalized computer-aided questions have a positive effect on their own learning progress?
3. Do the personalized computer-aided questions have a positive effect on student learning achievement, compared with the traditional computer-aided questions?
4. How can personalized questions be selected in order to provide the appropriate instructional support?

## 2 RELATED WORK

Sections 2.1 and 2.2 will respectively address the two following questions: What is the advantage of personalized learning and adaptive assessment? How has computer-aided question generation developed to help language assessment?

### 2.1 Personalized learning and Computerized Adaptive Test

Contemporary learning theories emphasize the importance of providing appropriate instructional support and guidance in order to facilitate meaningful learning and maximize students' learning outcomes. According to [27], for example, one of the important roles of education is to advance learning by providing appropriate instructional scaffolding to learners in their Zone of Proximal Development (ZPD), which refers to the difference between what a learner can do with and without an external help [28]; and closely related to the concept of ZPD is the notion that that appropriate support during the learning process can optimize learner achievement in their learning goals. Many education researchers have suggested that appropriate instructional scaffolding should at first be provided to help learners acquire knowledge, and then taper off when it becomes unnecessary [29], [30]. For instance, Chen et al. [2] considered a learner's ability for recommending personalized learning paths in a Web-based programming learning system, while Chen and Chung [3] analyzed students' understanding by suggest-ing English vocabulary on mobile devices. Similarly, within adaptive testing or practice, Barla et al. [5] , calculated an examinee's ability based on Item Response Theory [31] to select suitable questions, while Klinkenberg et al. [8] estimated a learner's ability and a question's difficulty based on the Elo [32] rating system for monitoring arithmetic in primary education. All of these studies demonstrated that personalized learning can improve student performance.

Computerized Adaptive Testing (CAT) [31] successively selects questions that adapt to an examinee's ability in order to maximize the precision of an examinee's ability. However, before the adaptive administration, the CAT requires a large human-written item bank and needs a sizable sample of people to pre-administer and trial the questions in order to calibrate the parameters of questions, e.g. difficulty and discrimination. When the questions and parameters are available, CAT presents the first question with a median difficulty to be administered, and then scores the response and estimates the examinee's ability. In the next iteration, CAT selects the question with the most information in terms of the current examinee's ability. CAT repeatedly administers questions and updates the examinee's ability until a certain stopping criterion is met, e.g. the examinee's standard error falls below a specific value. For most examinees' perspective, CAT provides a more precise understanding and saves about half the testing time. However, the prerequisite of CAT is the requirement of manual effort to generate the questions and calibrate the characteristics of the questions. Generally, during item calibration, an item should be answered by a large number of people, ideally between 200 to 1000 people, in order to estimate reliable parameters for the items [33], [34]. This procedure is very costly and time-consuming, and also less beneficial for online learning environments. It is especially impractical because the calibration had to be conducted repeatedly in order to get accurate norm referenced item parameters.

### 2.2 Computer-aided question generation in Language Learning

Computer-aided question generation is the task of automatically generating questions, which consists of a stem, a correct answer and distractors, when given a text. These generated questions can be used as an efficient tool for measurement and diagnostics. The first computer-aided question generation was proposed by [14]. Multiple-choice questions were automatically generated using three components: term extraction, distractor selection and question generation. First, noun phrases were extracted as answer candidates and sorted by term extraction. The more frequently that a terms appears, the more important the term becomes. The terms with higher term frequency consequently served as answers to the generated questions. Next, WordNet [35] was consulted by the distractor selection in order to capture the semantic relation between each incorrect choice and the correct answer. Finally, the generated questions were formed by predefined syntactic templates. Most of the following studies are based on such a system architecture.

An expanding body of research is now available that sheds light on the domain of English language learning, such as vocabulary, grammar and comprehension. This is because in these types of question generation, linguistic characteristics are analyzed to help produce items, just as experts do. In vocabulary assessment, Liu et al. [16] investigated word sense disambiguation to generate vocabulary questions in terms of a specific word sense, and considered the background knowledge of first language of examinees to select distractors. [18] analyzed the semantics of words and developed an algorithm to select candidates as a substitute word from WordNet [35], which were filtered by web corpus searching; they presented adjective–noun pair questions, including collocation, antonym, synonym and similar word questions in order to test students' semantic understanding. Turney [25] used a standard supervised machine learning approach with feature vectors based on the frequencies of patterns in a large corpus to automatically recognize analogies, synonyms, antonyms, and associations between words, and then transformed those word pairs into multiple–choice SAT (Scholastic Assessment Tests) analogy questions, TOEFL synonym questions and ESL (English as second language) synonym–antonym questions.

In grammar assessment, Chen et al. [15] focused on automatic grammar quiz generation. Their FAST system analyzed items from the TOEFL test and collected documents from Wikipedia to generate grammar questions using a part–of–speech tagger and predefined templates. Lee and Seneff [19] particularly discussed an algorithm to generate questions for prepositions in language learning. They proposed two novel distractor selections, one was applied a collocation–based method and the other was the usage of the deletion error in a non-native corpus.

In a reading comprehension assessment, the MARCT system [17] designed three question types: true-false question, numerical information question and not-in-the-list questions. In the true-false question generation, the authors replaced words in a sentence, extracted from an article on the Internet, with the synonyms or antonyms by using WordNet [35]. In the numerical information question generation, they listed some specific trigger words, such as "kilogram", "square foot", and "foot", corresponding to some predefined templates, such as "what is the weight of", "how large", and "how tall". In the not-in-the-list question generation, they used terms listed in Google Sets to identify the question type and select distractors. Differing from previous methods, Mostow and Jang [20] designed different types of distractor to diagnose the cause of comprehension failure, including ungrammatical, nonsensical, and plausible failures. The plausible distractors especially considered the context in the reading materials; they used a Naïve Bayes formula to score the relevance to the context in paragraph and words earlier in sentence. A student's comprehension was thus judged not only by evaluating one's vocabulary knowledge but also by testing the ability to decide which word is consistent with the surrounding context.

Even though the previous studies in the field of computer-aided question generation automatically generated all possible questions based on their proposed approach in an attempt to reduce the cost of time and money of manual question generation, such an exhaustive list of questions remains inappropriate for language assessment, because it can lead to redundant and overly–simplistic test questions that are unsuitable for evaluating student progress.

## 3 PERSONALIZED COMPUTER-AIDED QUESTION GENERATION

In this section, we will describe the constraints on vocabulary questions, grammar questions, and reading comprehension questions in an effort to answer the following question: How to generate personalized questions in different question types? Table 1 summarizes how to define the question difficulties and how distractors are selected, and Table 2 shows the four question types as examples, where the bolded words represent the stems, the bold italics the answers (also called target words in this study), and the other plausible choices in the questions are the distractors.

### 3.1 Vocabulary questions

The difficulty of a vocabulary question is determined by the difficulty of the correct answer. We assume that if a student selects the correct answer, he/she probably understands the question stem and distinguishes the correct answer from the distractors. Here, the difficulty of a word refers to word acquisition, the temporal process by which learners learn the meaning, understanding and usage of new words. For most learners of English as a foreign language, the acquisition age distribution of different words can be drawn from the inference from textbooks or a word list made by experts, because they learn foreign language depending on materials they study, not the environment they live in. In this study, the word difficulty is determined by a graded word list made by an education organization, the College Entrance Examination Center (CEEC) of Taiwan (http://www.ceec.edu.tw/research/paper_doc/ce37/5.pdf ). It contains 6,480 words in English and is divided into six levels, each of which represents the grade in which a word should be taught, as the a form of acquisition age. For each word from the given text, we identify its difficulty by first referencing its difficulty level from within the word list. When given the vocabulary proficiency level of a student, words with the same difficulty level in the given reading material were selected as the basis to form test questions. In the distractor selection, we also consulted the same graded word list as the source of distractor candidates. The distractors were selected according to the following criteria: word difficulty, part-of-speech (POS), word length and Levenshtein distance.

TABLE 1
DESIGN OF PERSONALIZED QUESTIONS WITH DIFFERENT QUESTION TYPES: VOCABULARY, GRAMMAR, AND READING COMPREHENSION QUESTIONS.

| | Vocabulary question | Grammar question | Independent referential question | Overall referential question |
|---|---|---|---|---|
| **How to define question difficulty?** | a graded word list | grammar frequency | reading difficulty estimation | |
| **How to select a target sentence (with answer)?** | a word | a sentence | a referent (not a singleton) | |
| **Stem template** | In the sentence "… ______ …", the blank can be: | In the Sentence, "… ______ …", the blank can be filled in: | The word "[*target word*]" in this sentence "[*target sentence*]" refer to: | Which of the following statements is TRUE? |
| **Distractor candidate source** | words from a graded word list | grammar patterns defined by a grammar book | other noun phrases (common nouns or proper nouns) in the given article | |
| **Distractor selection** | word difficulty | disambiguation | non-anaphora | |
| | part-of-speech | | not pronoun | |
| | word length | | number | |
| | Levenshtein distance | | gender | |

TABLE 2
A PARAGRAPH AND EXAMPLE GENERATED QUESTIONS.

| **Halloween** |
|---|
| Halloween, which falls on October 31, is one of the most unusual and fun holidays in the United States. It is also one of the scariest! **It is <u>associated</u> with ghosts, skeletons, witches, and other scary images.** …**Many of the original Halloween traditions <u>have developed</u> today into fun activities for children.** The most popular one is "trick or treat." On Halloween night, <u>children</u> dress up in costumes and go to visit their <u>neighbors</u>. When someone answers the door, the children cry out, "trick or treat!" What this means is, "Give us a <u>treat</u>, or we'll play a <u>trick</u> on you!" …This tradition comes from an old Irish story about a man₁ named **Jack₂** who was very stingy. ... But he₃ also could not enter hell, because he₄ had once played a trick on the <u>devil₅</u>. **All <u>he</u> could do was walk the earth as a ghost, carrying a lantern**… |

1. **In the sentence "It is __________ with ghosts, skeletons, witches, and other scary images.", the blank can be:**
   (1) distributed (2) *associated* (3) contributed (4) illustrated
2. **In the Sentence, "Many of the original Halloween traditions __________ today into fun activities for children.", the blank can be filled in:**
   (1) *have developed* (2) have developing (3) is developed (4) develop
3. **The word "he" in this sentence "All he could do was walk the earth as a ghost, carrying a lantern" refer to:**
   (1) ghost (2) devil (3) witch (4) *Jack*
4. **Which of the following statement is TRUE?**
   (1) On Halloween night, neighbors dress up in costumes and go to visit their children.
   (2) What this means is, "Give us a trick, or we'll play a treat on you!"
   (3) But the devil also could not enter hell, because he had once played a trick on the witch.
   (4) *Jack was so stingy that he could not enter heaven when he died.*

The first question in Table 2 is a vocabulary question. When a certain knowledge level of a student is known, the corresponding difficulty level of words, e.g. "associate", are identified from the graded word list. The sentence containing the word, "It is associated with ghosts, skeletons, witches, and other scary images", is then extracted to form a question and take "associate" as the correct answer. We also consult the same word list to select distractors which have the same difficulty level and the same part-of-speech (verb), and the least small distance of word length (9) and the least small Levenshtein distance among distractor candidates (distributed:6, illustrated:7, contributed:7).

## 3.2 Grammar questions

The difficulty of a grammar question, which is similar to that of a vocabulary question, is determined by the difficulty of the grammar pattern of the correct answer. Unfortunately, unlike the aforementioned word list, there is no predefined grammar difficulty measure available. In addition, second language learners usually learn grammati-cal structures simultaneously and incrementally, while native speakers have learn almost all grammar rules before formal education. Second language learning materials are often organized according to a well-thought learning plan. Thus, we assigned the difficulty of a grammar pattern based on the grade level of the textbook, which represents the age of grammar acquisition.

The difficulties of grammar patterns rely on the grade level of the textbook in which it frequently appears, which represents the age of grammar acquisition. We manually predefined 44 grammar patterns from a grammar textbook for Taiwanese high school students and automatically calculated the rate of occurrence of the grammar patterns in a set of English textbooks. First, we used the Stanford Parser [36] to produce constituent structure trees of sentences; next, Tregex [37], a searching tool for matching patterns in trees, was used to recognize the instances of the target grammar patterns in the set of textbooks. This corpus contains 342 articles written by different authors and collected from five different publishers (The National Institute for Compilation and Translation, Far East Book Company, Lungteng Cultural Company, San Min Book Company, and Nan-I Publishing Company).

To generate the grammar distractors, we also consulted the same grammar textbook and manually predefined distractor templates. These templates also needed to ensure that there were no ambiguous choices in the templates. Sometimes, it was possible that more than one grammar pattern could be a correct answer in a sentence. For example, the stem in the second question in Table 2, a distractor candidate "develop" could be consistent with the syntax of the target sentence regardless of the global context. Thus, we referred to the grammar textbook and an expert for designing distractor templates for each grammar pattern (examples shown in Table 3) in order to ensure the disambiguation of the distractors.

The second question in Table 2 is a grammar question. The target testing purpose in the second question is "present perfect tense", which is taught in a certain grade level. The original sentence is "Many of the original Halloween traditions have developed today into fun activities for children". The parse structure of the original sentence is depicted in Fig. 1. The grammar pattern of this parse

## TABLE 3
### EXAMPLES FROM THE DISTRACTOR TEMPLATES.

| level | function name | example answer | example distractor 1 | example distractor 2 | example distractor 3 |
|---|---|---|---|---|---|
| 1 | Perfect Tense | has grown | have growing | have been grown | had grown |
| 2 | Adjective (too...to) | too happy to | too happy that | too happiest to | none of the above |
| 3 | Prep Ving | in helping | in being help | in helped | in being helping |
| 4 | Gerund as Object | avoid taking | avoid to taking | avoid to take | avoid to took |
| 5 | Passive | is used | is using | used | will be using |
| 6 | Remember | remember to take | remembering to take | remember to taking | none of the above |

structure can be automatically identified by the Tregex pattern: /S.?/ < (VP < (/VB.?/ << have|has|haven't|hasn't)): /S.?/ < (VP < (VP < VBN)). When a grammar pattern is recognized, the difficulty degree of the grammar question is assigned based on the matched grammar pattern.

## 3.3 Reading Comprehension questions

The difficulties of the reading comprehension questions were based on the reading levels of the reading materials themselves. We assume that an examinee correctly answers a reading comprehension question because he/she can understand the whole story. The difficulty level of an article is correlated to the interaction between the lexical, syntactic and semantic relations of the text and the reader's cognitive aptitudes. The reading level estimation of a given document in recent years has increased noticeably. However, most of the past literature targeted first language learners, but the learning timeline and processing between first language learners and second language learners is different. In this study, we adopted the measure of reading difficulty estimation [38] designed for English as foreign language learners to identify the difficulty of reading materials, as a difficulty measure for the reading comprehension questions.

Reading Comprehension relies on a set of highly complicated cognitive processes [39]. In these processes, it is necessary to design an anaphora resolution, construction-integration model and to build a coherent knowledge representation [40]. Thus, in this work, we focus on the relation between sentences to generate two kinds of meaningful reading questions based on noun phrase co-reference resolution. Similar to Mitkov and Ha [14], who extracted nouns and noun phrases as important terms in reading material, we also focus on the interaction of noun phrases for testing purposes. The purpose of noun phrase co-reference resolution is to determine whether two expressions refer to the same entity in real life. An example is excerpted from Table 2 (This tradition…on the devil$_5$). It is easy to see that "Jack$_2$" means "man$_1$" because of the semantic relationship between the sentences. The follow-ing "he₃" and "he₄" are more difficult to judge as referring to "Jack₁" or "devil₅" when examinees do not clearly understand the meaning of the context in the document. This information is used in this work to generate reading comprehension questions, in order to examine whether learners really understand the relationship between nouns in the given context. In this work, the coreferential relations are identified by the coreference system [41].

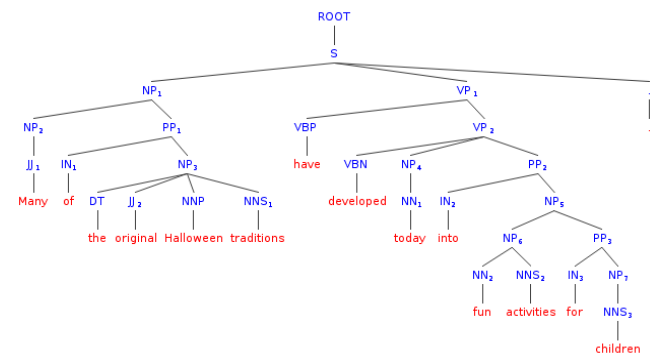

Fig. 1. The parse structure of the sentence "Many of the original Halloween traditions have developed today into fun activities for children".

There are two question types generated in the reading comprehension questions: an independent referential question for the single concept test purpose and an overall referential question for the overall comprehension test purpose. When a noun phrase is selected as a target word in the stem question, it should have an anaphoric relation with the other noun phrase. In the first type, a noun phrase (a pronoun, a common noun or a proper noun) is selected as a target word in the stem question, a noun phrase (a common noun or a proper noun) with the same anaphoric relation will be chosen as the correct answer and other noun phrases (common nouns or proper nouns) will be determined as the distractors. In the second type, the same technique for question generation applies to the sentence level. We regenerate new sentences as choices by replacing a noun (a pronoun, a common noun or a proper noun) with an anaphoric noun (a common noun or a proper noun) as the correct answer and substituting a noun with a non-anaphoric noun as distractors.

The third type of question in Table 2 is the independent referential question, which assesses one's understanding of the concept of an entity involved in sentences. The word "he" in the original sentence "All he could … a lantern" refers to "Jack", the distractors "ghost", "devil", and "witch" have non-anaphoric relations, are not pronouns, and are "singular" and "neutral". The fourth question type in Table 2 is the overall referential question, which contains more than one concept that needs to be understood. The correct answer is from the sentence "He was so stingy … died," and the word "He" is replaced with "Jack" because they have a referential relation. One of distractors is from "But he also could not … devil," where the word "he" refers to "Jack" instead of "devil", but we replace it with the non-anaphoric noun as a distractor. This approach further examines in the connection of concepts in the given learning material.

### 3.4 Personalized Quiz Strategy

In this section, we investigate how to select suitable questions as a quiz. It is critical to determine how to best form a test from a series of questions which match a student's proficiency level. In this study, a test was composed of fit questions, history-based questions and challenging questions. The fit questions indicate that a question's difficulty level is equal to a student's proficiency level, the history-based questions refer to a question's difficulty level that is easier than the student's proficiency level, and the challenging questions show that a question's difficulty level is more difficult than the student's proficiency level. The purpose of this test was not only to measure a student's proficiency level but also to review the previous relevant knowledge and stimulate future lessons. As Fig. 2 shows, we defined probability values to assign questions in a test. Similar to the normal distribution, the percentage of the history-based questions, the fit questions and the challenging questions were 20%, 60% and 20%, respectively. Moreover, since we wanted to correct the students' previous mistakes, when the fit questions were answered incorrectly, they were stored in the system. And so during the next iteration of the test, if there was any similar question based on the same concept, such as the present perfect tense, a question with this tense would be selected first. The goal was to enhance the students' knowledge and improve their proficiency.

The next question to be addressed is how to identify an examinee's proficiency level. Although much work has already been done, we still have two concerns which to date have attracted little attention. First, though every exercise performed by a student is recorded in most of the Web-based learning environments, most of the ability estimations only consider test responses at the time of testing, rather than incorporating a testing history. Second, the result of estimating an examinee's ability is often defined as a norm referenced value. It is thus hard to make a connection between an examinee's proficiency level and the difficulty level of a question. For these reasons, in this study we consider the ability estimation [42] which is determined by not only the current responses but also the testing history. Additionally, the estimated proficiency level corresponds to an explicit grade level in a school, rather than the implicit scale.

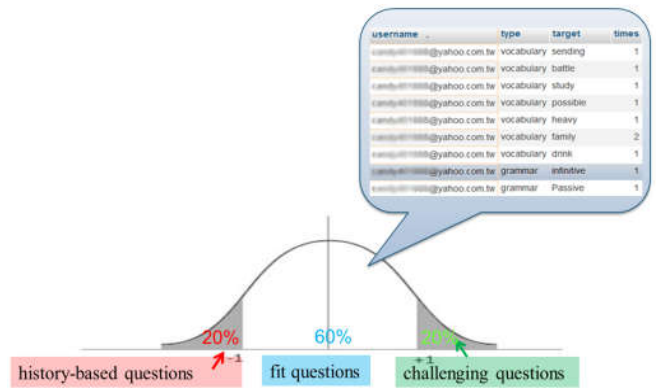

Fig. 2. The distribution of the selected questions.

## 4 EXPERIMENT

Sections 4.1 and Section 4.2 will respectively show how to evaluate the hypotheses and results of the following two questions: Do the personalized computer-aided questions help students correct their previous mistakes? (RQ1) Do the personalized computer-aided questions have a positive effect on their own learning progress (RQ2) and learning achievement compared to the traditional ones (RQ3)?

### 4.1 Experimental Design

The proposed methods were developed from the Au-toQuiz Project. The project was implemented on the IWiLL learning platform [43], which offered learners an online English reading and writing environment. The proficiency level in this study was defined from one to six, corresponding to the six semesters of the senior high school in Taiwan. A total of 2,481 questions with difficulties (from one to six) were automatically generated based on 72 new stories as reading materials. The news articles were collected from several global and local online news websites: Time For Kids, Student Times, Voice of America, CNN, China Post Online and Yahoo! News.

The experiment was held from July 1st to September 30th, 2011. The participants in this study were high school students in Taiwan, divided into two groups: a control group with general automatic quiz generation, and an experimental group with personalized automatic quiz generation. A total of 21 and 72 subjects in the control group and the experimental group respectively participated in the experiment. However, some students could not complete all phases for their personal reasons because the experiment was conducted during a summer vacation.

As Fig. 3 illustrates, the subjects during the experiment were asked to participate in twelve activities, consisting of reading an article and then taking a test. After the passage is presented and read, it is taken away so that the examinee cannot check the original sentences. Each test was composed of ten vocabulary questions, five grammar questions, and three reading comprehension questions. In addition, there was a pretest and a post-test for evaluating changes in learner proficiency, each with a similar degree of difficulty.

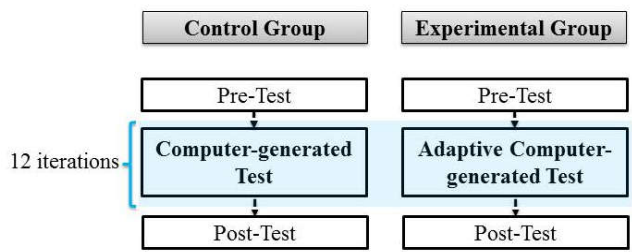

Fig. 3. Experimental procedure.

## 4.2 Experimental Results

### 4.2.1 The correction of previous mistakes

One of the aims of the personalized computer-aided questions was to help students correct their previous mistakes. We measured the rate at which students successfully corrected their mistakes on repeated concepts (denoted as the rectification rate) in the experimental group and control group, to determine the effect of generating items with repeated concepts. To make comparisons, the independent-samples t-test and the Mann-Whitney U test were both performed. Ideally, the distribution between the two groups is a normal distribution, making the t-test the appropriate choice. However, because of unequal sample sizes, the nonparametric method is complementary. The results suggest that the rectification rate in the experimental group was on average significantly higher than in the control group (t=6.60, p<0.001 in the independent-samples t-test and Z=-5.97, p<0.001 in the Mann-Whitney U test). Moreover, the subjects in the experimental group (M=0.54, SD=0.29) were more than half as likely to correct previous mistakes and answer similar questions (the same concepts but different questions) correctly. This indicates that the proposed personalized computer-aided questions would help the students figure out the unclear concepts.

### 4.2.2 Student performance

To further understand the influence of personalized automatic quiz generation, the normalized score (normalized from zero to one) in the post-test between the experimental group and control group were calculated and compared in both the parametric and nonparametric analyses. The results of an independent t-test (p=0.80 in the pretest and p=0.46 in the post-test) and the Mann-Whitney U test (p=0.99 in the pretest and p=0.59 in the post-test) showed no significant effect on the post-test between the experimental group and the control group. However, the paired sample t-test and the Wilcoxon signed-rank test showed a significant effect of the pretest and the post-test in the experimental group (p<0.01), while the performance of the control group had no statistically significant effect (p>0.05). This indicates that the personalized automatic quiz generation within the experimental group effectively improved their own learning.

To investigate how the students improved their own learning progress, we computed the percentage of correctly answered questions among the six difficulty levels in the pretest and the post-test. The tests comprised 28 items among six difficulty levels (six, three, six, three, seven and three questions per respective level, corresponding to levels one through six). A chi-square test for homogeneity of proportions was conducted to analyze the proportion between the pretest and post-test; Table 4 presents two contingency tables, one for the control group and the other for the experimental group. The results of the experimental group ($X^2(5)=16.24$, $p<0.01$) show significantly different proportions between the pretest and post-test, while the control group ($X^2(5)=7.46$, $p>0.05$) presented similar percentages among the six difficulty levels. This disparity reveals that the personalized test affected the ability of the students in the experimental group. To further investigate the difference in the experimental group, an a posteriori comparison revealed that the number of correctly answered questions with level two and level six in the post-test were statistically higher than those in the pretest, whereas the number of questions with level one and level four in the post-test were significantly lower than those in the pretest. This suggests that the number questions with a higher difficulty level that were correctly answered increased after the personalized quiz strategy.

TABLE 4
CONTINGENCY TABLES FOR THE NUMBER OF CORRECTLY AN-
SWERED QUESTIONS PER DIFFICULTY LEVEL IN THE PRETEST
AND POST-TEST.

| difficulty level | | 1 | 2 | 3 | 4 | 5 | 6 |
|---|---|---|---|---|---|---|---|
| Control group | Pretest | 23.8% | 9.3% | 21.7% | 12.4% | 23.4% | 9.3% |
| | Post-test | 21.3% | 14.6% | 21.0% | 9.6% | 20.7% | 12.8% |
| Experimental group | Pretest | **24.8%** | 9.9% | 20.9% | **12.9%** | 20.6% | 10.8% |
| | Post-test | 20.5% | **13.1%** | 22.6% | 9.5% | 21.1% | **12.7%** |

## 5  DISCUSSION

Sections 5.1 and 5.2 will discuss how to select personalized questions in order to provide the appropriate instructional supports (RQ4), and what the difference is between the proposed personalized learning environment and other previous ones.

### 5.1  The personalized strategy

Our experiment results show that the students could not only correct more similar questions which they answered incorrectly before but also correctly answer more difficult questions, when the difficulty of the learning material was adapted to their proficiency levels. Although they only made progress between the pre–test and post–test and there was no statistical difference on the post-test between the experimental group and the control group, this might be due to the difference in objectives between the test environments and learning environments. While the objective of the adaptive test environment is to measure a student's ability as precisely as possible, that of the adaptive learning environments is to optimize their learning efficiency [44] as much as possible. Students in Barla et al.'s [5] adaptive test environment, for example, were administered questions adjusted to their abilities because when the difficulty of a question is equal to the knowledge level of a student, it is optimally informative [31]. On the other hand, in the Computerized Adaptive Practice (CAP) system proposed by Klinkenberg et al. [8], easier questions were presented in order to increase learner motivation.

To investigate how to select effective questions with the proposed personalized automatic question generation, an experiment was held from August 4th to September 11th, 2012. The participants in this study were second-year high school students in Taiwan, divided into three groups: a proposed group for which questions were selected by the proposed quiz strategy (N=28), a motivated group for which questions were selected by matching the difficulty of questions one level lower than their proficiency level (N=40), and a challenged group for which questions were selected by matching the difficulty of questions one level higher than their proficiency level (N=30). The other setting in this comparison was the same in Section 4.1.

To evaluate the hypothesis that student performance improves more under either a motivating or a challenging condition as opposed to the proposed personalizing condition, the normalized score (normalized from zero to one) in the pretest and post-test among the three groups were calculated and compared. Table 5 lists the descriptive statistic results. An independent t-test of overall scores indicated a significant difference (p=0.35 in the pretest and p=0.04 in the post-test) between the proposed group and the motivated group, while there was no significant effect (p=0.38 in the pretest and p=0.56 in the post-test) between the proposed group and the challenged group. In addition, the scores for the grammar questions (p=0.07) and the reading comprehension questions (p=0.01) on the post-test revealed especially significant results between the proposed group and the motivated group. This suggests that the grammar and reading comprehension questions selected by matching the difficulty of questions one level easier than the students' proficiency level were able to motivate students and facilitate significant progress. This result is consistent with the findings of Klinkenberg et al. [8]. It therefore appears that an effective personalized instructional support of the proposed system should offer vocabulary questions at the same level of a student's proficiency level, and grammar and reading comprehension questions lower than the student's level.

### 5.2  The comparison of different test environments

To our knowledge, this work is the first empirical study to analyze student performance with automatically generated questions and a personalized test strategy. Table 6 provides a comparison of the proposed system with previous test environments. In the traditional test environment, examinees in the same grade or class usually take the same tests, which were previously made by experts. In the adaptive test environment (e.g. Barla et al., [5] and Klinkenberg et al., [8]), tests are likewise made beforehand by experts, but examinees in the same grade or class could have different tests depending on their ability. Tests based on computer-aided question generation (e.g. Mitkov & Ha, [14]) save both time and production costs, but they are usually not designed for any test purpose. In our method, questions and the difficulty of questions are not only generated automatically, but are also provided to examinees depending on various abilities and their previous mistakes. The examinee's performance is recorded in the system and the concepts behind incorrectly answered questions are reincorporated into future tests. Additionally, the estimated proficiency level in this study corresponds to an explicit grade level in a school, whereas that in the ability estimation of the traditional adaptive test environment is a point on an implicit scale. This retains the advantages of both the adaptive test environment and automatic question generation. That is, the proposed approach 1) offers students an effective approach to automatically measure their understanding, 2) clarifies their incorrect concepts, and 3) reduces the teacher's burden of question generation, allowing teachers to spend more time to teach and assist students.

TABLE 5
THE DESCRIPTIVE STATISTIC RESULTS (MEAN AND STANDARD DEVIATION) IN THE PRETEST AND POST-TEST AMONG THE THREE.

| group | vocabulary | | grammar | | reading comprehension | | overall | |
|---|---|---|---|---|---|---|---|---|
| | pretest | post-test | pretest | post-test | pretest | post-test | pretest | post-test |
| motivated | 0.52 (0.16) | 0.62 (0.16) | 0.57 (0.25) | 0.53 (0.20) | 0.41 (0.19) | 0.63 (0.24) | 0.50 (0.13) | 0.59 (0.14) |
| Proposed | 0.50 (0.24) | 0.64 (0.24) | 0.51 (0.32) | 0.45 (0.14) | 0.37 (0.25) | 0.47 (0.25) | 0.46 (0.22) | 0.52 (0.14) |
| challenged | 0.57 (0.21) | 0.65 (0.18) | 0.49 (0.20) | 0.46 (0.18) | 0.44 (0.22) | 0.52 (0.24) | 0.50 (0.16) | 0.54 (0.15) |

TABLE 6
COMPARISON OF DIFFERENT TEST ENVIRONMENTS.

| Comparison | Automation | Personalization |
|---|---|---|
| The traditional test environment | No | No |
| The adaptive test environment | No | Yes |
| The test environment using computer-aided question generation | Yes | No |
| The proposed personalized automatic question | Yes | Yes |

## 6 CONCLUSION

This work presents an adaptive test environment in order to enhance the understanding of English language learners. We proposed a personalized automatic quiz generation model to generate multiple–choice questions with varying levels of difficulty and selected questions depending on a student's estimated proficiency level as well as unclear concepts behind incorrect responses. We found that the students with our proposed method corrected more than 50 percent of the questions which they answered incorrectly, while students in the traditional test environment usually corrected only 10 percent of questions in a test. In other words, through the personalized learning and practice, the unclear concepts in different contexts are presented repeatedly so that students can correct their misunderstanding more frequently. Moreover, the students with the personalized automatic quiz generation not only corrected their errors but also answered more difficult questions correctly. These students demonstrated the most progress between the pre–test and post–test. It is likely that because they could correct their misunderstandings, they had enough ability for the challenging questions. Finally, we found that students could not only make significant progress but also sustain their learning motivation if the difficulty levels of the vocabulary questions were at the same level of a student's proficiency level, and if the difficulty of the grammar and reading comprehension questions was lower than the student's level.

Several implications can be drawn from this study if learners can in fact learn English with this system. First, it would provide a personalized learning environment. Students with different abilities could practice adaptive exercises with appropriate difficulties and be given the opportunity to repeatedly clarify unclear concepts. Second, it would provide teachers with diagnostic information to better understand students. Because students learn and practice on the platform, teachers could guide them for specifically difficult concepts. Moreover, the system records students' behavior, teachers could use this information to target weak areas of knowledge that students have. Finally, it could remove the barrier of physical academic textbooks; and by having online resources being made available and updated every day, learners would be able to learn something new anytime they want.

One of the limitations of our current research is that it was difficult to identify students' incorrect responses in reading comprehension questions, because we only defined a limited number of question types to generate questions. Although these questions were classified into various difficulties, they could still be insufficient to investigate students' understanding. One possible solution would be to observe and learn from the data; however, this would require researchers or students to label and define this resource. Another limitation is that this approach only focuses on English learning. The personalized framework may be applied to other language learning fields, but other disciplines, such as mathematics, the system would need to be redesigned. In our future work, we plan to develop a wide variety of quiz types, such as cloze tests or different reading comprehension items. We can imagine a scenario in which a second language learner reads up–to–date news and immediately takes a test to evaluate himself. Such tools will provide convenient language instruction and serve as a further complement to our existing testing system.

## ACKNOWLEDGMENT

This article combines material from [45] on the the preliminary results, and substantial new content including the evaluations and explainations. This work was partially supported by National Science Council, Taiwan, with Grant No. 100-2511-S-008-005-MY3 and NSC97-2221-E-001-014-MY3.